\documentclass[10pt,twocolumn,letterpaper]{article}

\usepackage{cvpr}              

\usepackage[accsupp]{axessibility}  

\usepackage{graphicx}
\usepackage{amsmath}
\usepackage{amssymb}
\usepackage{booktabs}
\usepackage{subcaption}  
\usepackage{multirow}  
\usepackage{array}  
\usepackage{booktabs}  

\DeclareMathOperator*{\argmin}{arg\,min}
\newcolumntype{P}[1]{>{\centering\arraybackslash}p{#1}}

\usepackage{hyperref}
\hypersetup{pagebackref,breaklinks,colorlinks}

\usepackage[capitalize]{cleveref}
\crefname{section}{Sec.}{Secs.}
\Crefname{section}{Section}{Sections}
\Crefname{table}{Table}{Tables}
\crefname{table}{Tab.}{Tabs.}

\def\cvprPaperID{***} 
\def\confName{CVPR}
\def\confYear{2023}

\begin{document}

\newcommand{\fig}[4]{\begin{figure}[!p] \centering
			\includegraphics[width=#2\columnwidth]{#1}
            \caption{\label{#3} #4}\end{figure}} 
\newcommand{\norm}[1]{{\left\lVert#1\right\rVert}^2}  
\newcommand{\normMAE}[1]{{\left\lVert#1\right\rVert}_{1}}  
\newcommand{\pa}[1]{\left( #1 \right)}  
\newcommand{\ba}[1]{\left[ #1 \right]}  
\newcommand{\bt}[1]{\textbf{#1}}  

\title{Recursions Are All You Need: Towards Efficient Deep Unfolding Networks}

\author{Rawwad Alhejaili\textsuperscript{1,2} , Motaz Alfarraj\textsuperscript{1,2}, Hamzah Luqman\textsuperscript{2,3} , and Ali Al-Shaikhi\textsuperscript{1}\\
\texttt{\{g202112950, motaz, hluqman, shaikhi\}@kfupm.edu.sa}\\
\textsuperscript{1} Electrical Engineering Department\\
\textsuperscript{2} SDAIA-KFUPM Joint Research Center for Artificial Intelligence\\
\textsuperscript{3} Information and Computer Science Department\\
King Fahd University of Petroleum and Minerals, Dhahran 31261, Saudi Arabia}

\maketitle

\begin{abstract}
    The use of deep unfolding networks in compressive sensing (CS) has seen wide success as they provide both simplicity and interpretability. However, since most deep unfolding networks are iterative, this incurs significant redundancies in the network. In this work, we propose a novel recursion-based framework to enhance the efficiency of deep unfolding models. First, recursions are used to effectively eliminate the redundancies in deep unfolding networks. Secondly, we randomize the number of recursions during training to decrease the overall training time. Finally, to effectively utilize the power of recursions, we introduce a learnable unit to modulate the features of the model based on both the total number of iterations and the current iteration index. To evaluate the proposed framework, we apply it to both ISTA-Net+ and COAST. Extensive testing shows that our proposed framework allows the network to cut down as much as 75\% of its learnable parameters while mostly maintaining its performance, and at the same time, it cuts around 21\% and 42\% from the training time for ISTA-Net+ and COAST respectively. Moreover, when presented with a limited training dataset, the recursive models match or even outperform their respective non-recursive baseline. 
    Codes and pretrained models are available at \url{https://github.com/Rawwad-Alhejaili/Recursions-Are-All-You-Need}.
\end{abstract}

\section{Introduction}

    \begin{figure*} 
        \centering \includegraphics[width=2\columnwidth]{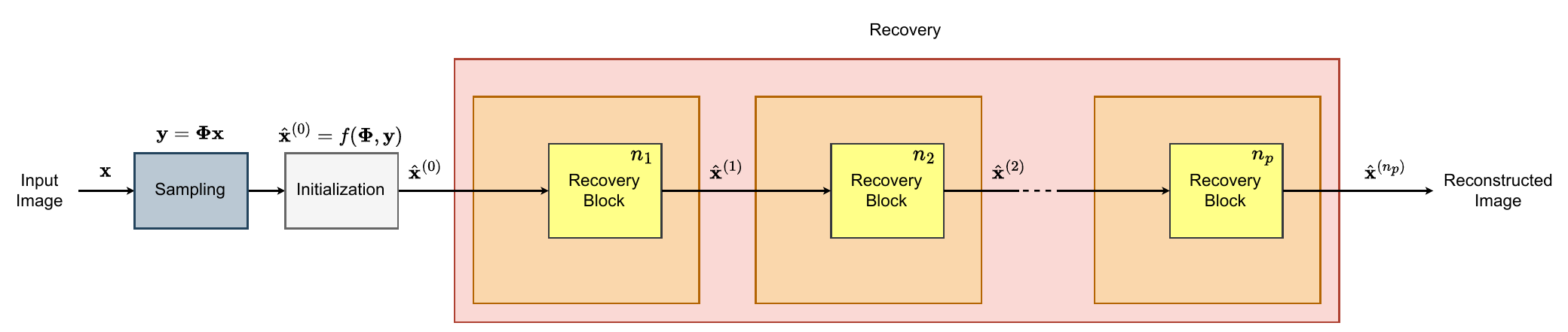}
        \caption{General structure of deep unfolding models.
        \label{cs_skeleton}
        }
    \end{figure*}

    \begin{figure*} 
        \centering \includegraphics[width=2\columnwidth]{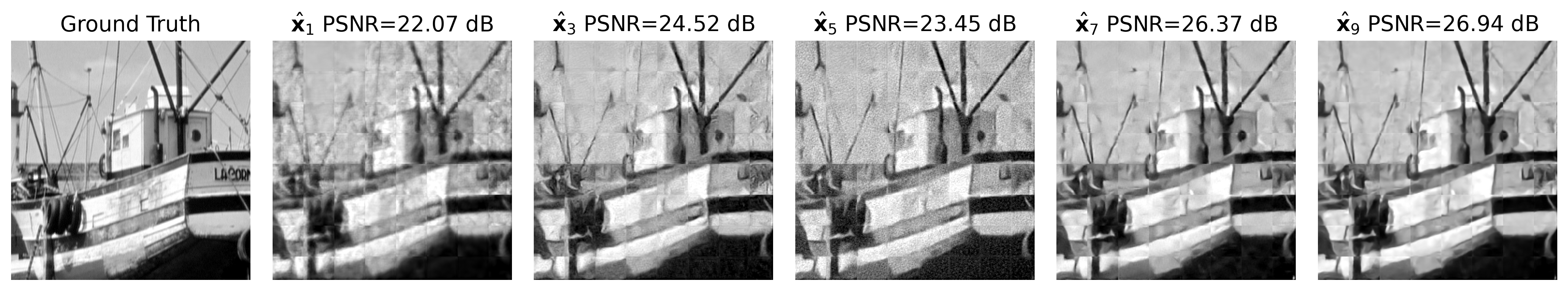}
        \caption{Intermediate CS reconstruction results of ISTA-Net \cite{ista-net}. 
        \label{ista-net phase2phase}
        }
    \end{figure*}
    
	Compressive sensing (CS) is an emerging field that challenges the conventions of digital data acquisition. The Nyquist sampling rate (the golden standard in signal acquisition) states that any signal $x$ can be recovered when it is sampled at double its highest frequency. However, sampling the signal at this rate can pose some challenges. For example, in data centers, the Nyquist rate limits the amount of data stored on storage drives leading to insufficient utilization of the servers. 
    Also, in seismic surveys, sampling the signal at the Nyquist rate is challenging either due to geological or economical constraints \cite{Prof_Wail_Book}. As such, while the Nyquist rate ensures perfect reconstruction of the signal, it can be unappealing in many cases. However, CS promises to recover signals below the Nyquist rate \cite{cs_book} \cite{cs_book_2013_math} \cite{cs_candes2006robust}.
    Mathematically, the recovery problem can be formulated as:
    \begin{align}
        \mathbf{y} = \mathbf{\Phi} \mathbf{x}
        \label{cs_problem}
    \end{align}
    where $\mathbf{y} \in \mathbb{R}^M$ is the measurements, $\mathbf{x} \in \mathbb{R}^N$ is the original signal, and $\mathbf{\Phi}  \in \mathbb{R}^{M\times N}$ is the sampling matrix.
    
    Compressive sensing aims to reconstruct the signal (or image) when the samples $M \ll N$. This can be guaranteed with high probability when the signal is sparse enough in one domain \cite{cs_book} \cite{cs_book_2013_math} \cite{cs_candes2006robust}. This is why CS emphasizes sparsity. Representing the collapsed signal is done by:
    \begin{align}
        \mathbf{y} = \mathbf{\Phi} \mathbf{\Psi} \mathbf{s}
    \end{align}
    where $\mathbf{\Psi} \in \mathbb{R}^{N \times N}$ is the sparsifying basis and $\mathbf{s}$ is the signal $\mathbf{x}$ in the sparse domain.
    
    Although compressive sensing presents theoretical guarantees about the recoverability of the signal \cite{cs_candes2006robust}, it is still a challenging problem to solve. Finding the sparsest solution requires iterative calculations which in turn makes CS computationally expensive to run \cite{omp} \cite{ista} \cite{lasso}.
    
    However, applying deep learning (DL) based approaches to compressive sensing proved to be fruitful. Neural networks are able to jointly learn the sampling matrix and the inverse mapping of the measurements $y$ to the original signal $x$ \cite{csnet}\cite{csnet_plus}\cite{opine} \cite{casnet2022}. Neural network-based CS methods demonstrated both better performance than handcrafted techniques and more importantly, at a lower computational complexity \cite{ista-net} \cite{csnet} \cite{coast2021}. 

    For deep learning-based CS models, deep unfolding networks are an attractive choice. Those models utilize the rich literature in the CS field by taking existing iterative handcrafted CS methods and supercharge them with the use of neural networks \cite{ista-net} \cite{admm-net2016} \cite{coast2021}. This is done by mapping each iteration to its own neural network block. The advantage this provides is both interpretability and simplicity.
    
    Deep unfolding models usually exhibit improved performance as the number of recovery blocks increases \cite{ista-net} \cite{coast2021}. However, we argue that this increase in performance could be attributed to the increased number of iterations rather than the increase in recovery blocks. Therefore, increasing the number of recovery blocks unnecessarily increases the number of parameters and redundancies. This makes the model more prone to overfitting necessitating a larger and more diverse training dataset \cite{overfitting_2019} and can also limit its deployment options \cite{model_compression_survey_2017}.

    In this work, we propose a recursive framework to increase the efficiency of deep unfolding networks in terms of training time, number of parameters, and training data efficiency. Specifically, our contribution is three-fold:
    \begin{itemize}
        \item First, we cut down on the redundancies in the deep unfolding networks by introducing recursions. This allows the model to increase its utilization of each layer's capacity before moving on to the next layer or block. 
        \item Secondly, we propose the use of random recursions during training to decrease the training time and allow the model to work with varying degrees of recursions.
        \item Finally, we acknowledge that a naive implementation of recursions may not be optimally efficient. Therefore, we leverage a learnable unit to modulate the features of the model based on the total number of iterations and the current iteration index.
    \end{itemize}
    
    The proposed framework is applied to both ISTA-Net+ \cite{ista-net} and COAST \cite{coast2021}. The results on COAST show that the proposed framework decreases the training time by 42\% and the learnable parameters by almost 75\% while sustaining a minimal impact on its performance. In the case of ISTA-Net+, we observe similar trends. The recursive framework decreases the training time by 21\% and the learnable parameters by 66\%. However, we find that the recursive ISTA-Net+ model slightly outperforms its baseline model. 
    Furthermore, when the size of the training data is limited, the recursive models either match their respective baselines (which is the case for COAST) or even outperform the baseline in the case of ISTA-Net+. Therefore, the recursive models are more resilient to overfitting thanks to the inherent reduction of their learning parameters.

\section{Literature Review}
    Previous work in compressive sensing can be categorized into two categories: handcrafted iterative optimization-based methods and data-driven deep learning-based methods. The data-driven methods can also be categorized into two categories, i.e., deep straightforward networks and deep unfolding networks.

	\begin{figure*}
		\centering \includegraphics[width=2\columnwidth]{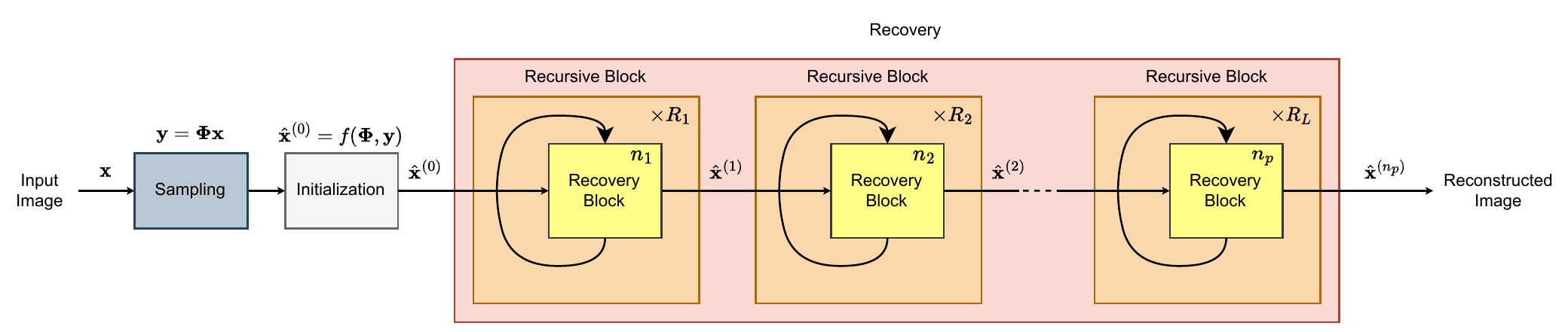}
		\caption{General architecture of the recursive framework. Compared to general deep unfolding models such as COAST \cite{coast2021} and ISTA-Net+ \cite{ista-net}, $R_i$ recursions are used in each recovery block $i$ in the recovery subnet.
			\label{recursive_framework}
		}
	\end{figure*}

\subsection{Iterative Optimization Based Methods}
    Generally, inverse problems such as in \cref{cs_problem} are impossible to solve as the number of unknowns $N$ is usually much greater than the number of measurements $M$. However, compressive sensing theory states that if the signal $x$ is sparse in some domain, then it can be recovered with high probability. Ideally, the recovered signal will have the lowest $l_0$-norm \cite{cs_book_2013_math}. Though, $l_0$-norm minimization is a non-convex optimization problem and is prone to combinatorial complexity \cite{cs_book_2013_math}. As such, computing the minimum $l_0$-norm suffers from a severe computational cost as the signal $x$ increases in size. Others have proposed minimizing the $l_1$-norm instead \cite{cs_candes2006robust}, which translates the problem into a convex optimization problem. Such methods include orthogonal matching pursuit (OMP) \cite{omp}, least absolute shrinkage and selector operator (LASSO) \cite{lasso}, 
    and iterative shrinkage-thresholding algorithm (ISTA) \cite{ista}. Other works have proposed minimizing the total variation (TV) instead, which include methods such as TVAL3 \cite{tval3} and denoising-based approximate message passing (D-AMP) \cite{d-amp}. Overall, while handcrafted methods have made strides in decreasing their associated computational complexity, they still remain computationally demanding due to their reliance on iterative optimizations.

\subsection{Deep Straightforward Models}
    Briefly, deep straightforward networks model the CS problem purely for neural networks and are not based on a handcrafted CS method.
    \cite{sda} proposes stacked denoising autoencoder (SDA), which uses multi-layer perceptrons (MLPs) for both the sampling and reconstruction stages. ReconNet \cite{reconnet} uses a random Gaussian matrix in the sampling stage and then uses convolutional neural networks (CNNs) for reconstruction followed by a conventional denoiser (BM3D \cite{bm3d}). Shi \textit{et al}. \cite{csnet} proposed CSNet. It jointly learns both the sampling matrix and the inverse mapping from the measurements to the original signal. Notably, it introduced the use of convolutions to learn the sampling matrix. Furthermore, other works \cite{csnet_plus} enforce constraints on the learned sampling matrices (such as limiting them to binary numbers), so that they can be applied more easily to hardware solutions. In \cite{csnet_plus}, they show that the learned sampling matrices can improve the reconstruction results even when applied to handcrafted methods.
    
    \begin{figure*}
    	\centering
    	\begin{subfigure}[b]{0.95\columnwidth}
    		\centering
    		\includegraphics[width=1\columnwidth]{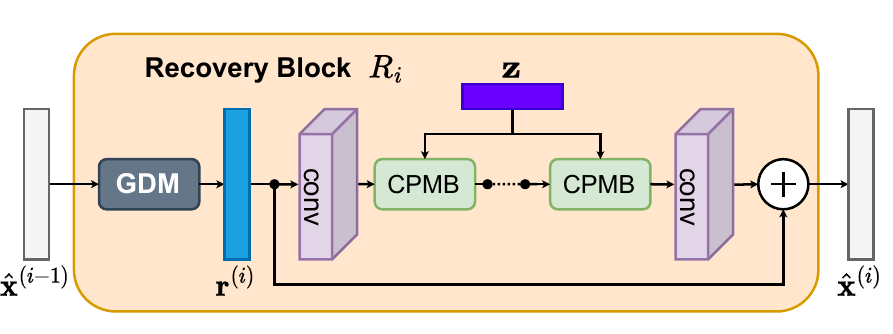}
			\caption{Recovery block of COAST \cite{coast2021}.}
			\label{coast_recovery}
    	\end{subfigure}
    	\hfill
    	\begin{subfigure}[b]{0.7\columnwidth}
    		\centering 
    		\includegraphics[width=1\columnwidth]{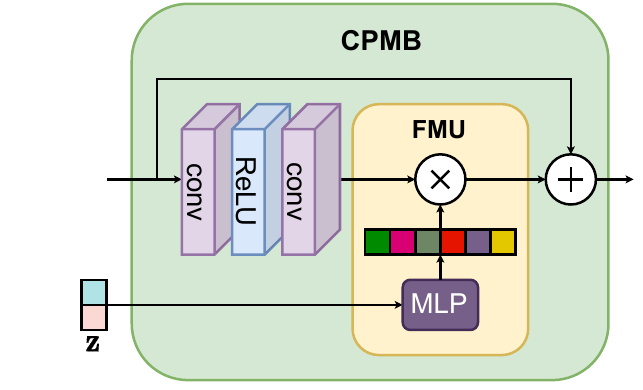}
    		\caption{CPMB block of COAST \cite{coast2021}.}
   			\label{CPMB}
    	\end{subfigure}
    	\hfill
    	\begin{subfigure}[b]{0.40\columnwidth}
    		\centering 
    		\includegraphics[width=0.55\columnwidth]{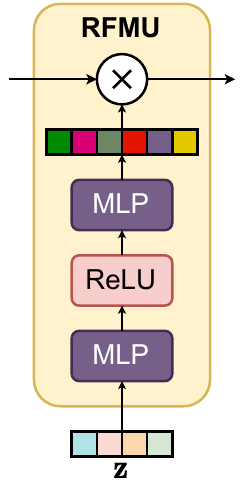}
    		\caption{The proposed RFMU unit.}
    		\label{fig:RFMU}
    	\end{subfigure}
    
    	\caption{Detailed look into the recovery block of COAST \cite{coast2021} and the RFMU unit. Note that the RFMU unit has two extra parameters in $\mathbf{z}$ compared to the FMU unit in COAST \cite{coast2021}, and they are the current recursion index $R_{\mathit{cur}}$ and the total number of iterations $R_{\mathit{tot}}$.}
    	\label{}
    \end{figure*}

    Many CS architectures suffer from being limited to a single CS ratio (and have to be retrained for each specific ratio). Wuzhen Shi \textit{et al}. proposes SCSNet \cite{Scalable_CS}, which solves the problem by adding enhancement layers. This allows the model to tackle multiple CS ratios without the need for retraining. A more recent approach by Vladislav Kravets \& Adrian Stern \cite{ProgressiveCS2022} solves this by progressively sampling the image at different scales where each scale builds on the measurements obtained from the last one. In other words, they reconstruct the low frequencies first and then progressively reconstruct higher frequencies. As such, the model can handle different CS ratios by design.

\subsection{Deep Unfolding Models}
    Deep unfolding models are neural networks that are designed with handcrafted iterative CS techniques as their basis. They map each iteration of the handcrafted technique to a neural network block which allow them to utilize the speed and performance of neural networks while preserving the interpretability of the handcrafted techniques. It also helps to keep the network design fairly simple.

    One such network is ISTA-Net \cite{ista-net}. As the name implies, the network revolves around the iterative shrinkage-thresholding algorithm (ISTA) \cite{ista}. The performance of ISTA highly depends on the sparsity of the input, so finding the best domain that ensures the highest sparsity is crucial for this algorithm. As such, ISTA-Net maps each iteration of ISTA to a corresponding block and utilizes the representative power of CNNs to find the sparsest possible transform. By default, the network uses 9 blocks (with an identical structure) to reconstruct the image. 
    However, the network requires training for each specific CS ratio meaning a single model cannot handle multiple CS ratios. Also, at low CS ratios, the network suffers from blocking artifacts due to its block-based recovery. OPINE-Net \cite{opine} modifies ISTA-Net by letting it learn its optimal sampling matrix.

    COAST \cite{coast2021} is another deep unfolding network that is built on top of ISTA-Net. It solves the main two issues with ISTA-Net. First, to promote the network to recover the image from different CS ratios, they introduce random projection augmentation (RPA), which exposes the network during training to many different sampling matrices at different CS ratios. Furthermore, they add a controllable unit that modulates the features based on the CS ratio. Secondly, they address the blockification issue of ISTA-Net by introducing a plug-and-play deblocking module that deblockifies the features before feeding them to the CNN layers. On top of solving the issues of ISTA-Net, the model also introduces changes to the recovery stage of ISTA-Net for optimal recovery, and instead of using 9 blocks for recovery, the model now uses 20 blocks (with identical structure).
	
	Regarding recursions specifically, there have been works that used recursions to decrease the number of learnable parameters. DPDNN \cite{dpdnn} is one such network. It uses an encoder-decoder network in the recovery block to reconstruct the image over multiple blocks. However, all blocks share the same learning parameters, and thus, it can be described by \cref{recursive_framework} as having a single recovery block but with $R_1=6$. The authors of ISTA-Net \cite{ista-net} have also tested sharing the learnable parameters across all recovery blocks, which can be described by \cref{recursive_framework} as having a single recovery block and $R_1=9$, and they found that the model still performed competently despite the significant decrease in the number of parameters. 
 
    With that said, the naive implementation of recursions leaves a lot to be desired. For one, the previous methods used a fixed number of recursions during training, and thus, did not enjoy any reduction to the training time. Secondly, the features of the model were not modulated to account for recursions. This is a crucial point as different features need to be emphasized depending on the iteration index and will enable efficient use out of recursions. Finally, recursions on those models were applied when the network only had a single recovery block, thus, its recovery potential was limited to the capacity of that single block.

    Now, we move on to what would be our major contribution in this work. From \cref{cs_skeleton}, we see that in general, deep unfolding models reconstruct the image iteratively over multiple blocks but with identical structures. \cref{ista-net phase2phase} shows how such an iterative model can improve the results in each phase or block, and how in the specific case of ISTA-Net \cite{ista-net}, the model can possibly improve its results with extra phases. However, this approach is unattractive as it increases the number of parameters, and as such, runs the risk of vanishing gradients and overfitting, which is especially unappealing when the size of the training dataset is small \cite{overfitting_2019}.

    To reduce the redundancy in the recovery phases, we propose the use of recursions. The intuition behind this is instead of passing the reconstructed signal to the next recovery block in each iteration, we will feed it back to the same block for $R_{i}$ iterations until the point of diminishing returns (ideally). This should provide us with three major advantages.
    \begin{itemize}
        \item First, we would be able to realize more of the capacity of each recovery block, allowing for more efficient use of them, which will lead to a lower number of trainable parameters.
        \item Secondly, the training time can be decreased by randomizing the number of recursions per recovery block in each training iteration.
        \item Thirdly, due to the decrease in the number of parameters, recursive models are more resilient to overfitting and thus, can perform better when the training dataset is small.
    \end{itemize}

    Finally, do note that the use of recursions should not be limited to compressed sensing applications. We believe that it might be generalizable to most image restoration tasks such as reconstruction, denoising, and interpolation as long as both the input and the output are in the same domain. Deep unfolding models lend themselves especially well for recursions due to their iterative design, which usually ensures that both the input and the output remain in the same domain.

    \begin{figure*}
		\centering \includegraphics[width=2\columnwidth]{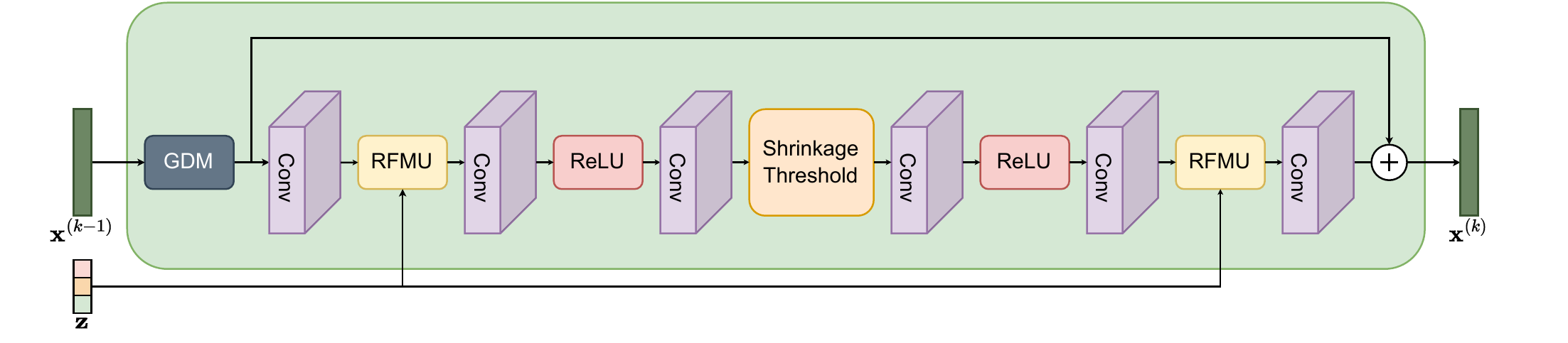}
		\caption{Recovery block of ISTA-Net+ \cite{ista-net} after the addition of the RFMU unit. 
			\label{ISTAP_RFMU}
		}
	\end{figure*}

\section{Methodology}
    COAST \cite{coast2021} will be used to demonstrate how the recursive framework can be applied to a deep unfolding network. We will discuss the general details of COAST and how the recursive framework was applied to it. Then, we will discuss how the framework modulates the features of the network based on recursions. Finally, we discuss how we modulate the features of the network based on recursions.

\subsection{COAST}
    The COAST model follows the general structure outlined in \cref{cs_skeleton}. The image is sampled using random Gaussian matrices with orthogonal rows, and the initialization is done by:
    \begin{align}
        \hat{\mathbf{x}}^{(0)} = \mathbf{\Phi}^{T} \mathbf{y}
        \label{init}
    \end{align}
    Afterward, the initial solution is fed to the recovery network.
    
    The fundamental recovery block of COAST is illustrated in \cref{coast_recovery}. First, the image is fed to a gradient descent module (GDM) to preserve the ISTA \cite{ista} structure. It is computed by:
    \begin{align}
        \mathbf{r}^{i} = \hat{\mathbf{x}}^{(i-1)} - \rho^{(i)}
                         \mathbf{\Phi}^{T} (\mathbf{\Phi}\hat{\mathbf{x}}^{(i-1)} - \mathbf{y})
    \end{align}
    where $\rho$ is the learning rate. Next, the image is fed to a convolutional layer to project it to the feature space with $C=N_{f}$. Then, a stack of controllable proximal mapping blocks (CPMB) is used to solve the proximal mapping problem of ISTA, which is defined as:
    \begin{align}
        \textbf{prox}_{\lambda \psi}(\mathbf{r}) = \argmin_{\hat{\mathbf{x}}} \frac{1}{2} \norm{\hat{\mathbf{x}} - \mathbf{r}}_2 + \lambda \psi (\hat{\mathbf{x}})
    \end{align}
    where $\lambda$ is a regularization parameter, and $\psi(\cdot)$ is a transformation function that is now learned by the CPMB block. Furthermore, to let the model better adapt to the different CS ratios and noise levels, \cite{coast2021} proposes a feature modulation unit (FMU) that modulates the features of the CPMB block based on the CS ratio and noise level. Specifically, the model employs a linear layer that takes those two parameters and outputs $N_f$ modulation coefficients where each one is used to modulate its respective feature in the CPMB block. Finally, a convolutional layer is used to project the image from the feature space back to the spatial domain.
\subsection{Recursions}
    Since COAST \cite{coast2021} follows the general structure of deep unfolding models (shown in \cref{cs_skeleton}), introducing recursions to it is fairly straightforward. A feedback connection is added from the output of each recovery phase back to its input, and $R_i$ denotes the number of iterations per layer (IPL). The modified network will now have the same structure seen in \cref{recursive_framework}.

    To further promote the model to learn recursions, we will introduce random recursions. In each training iteration, we will randomize the IPL of each layer (with $R_i$ being the highest number of IPL and 1 being the lowest). This will allow the model to generalize better to a wide range of recursions instead of being trained for a single set of recursions. Additionally, it will decrease the training time since on average fewer FLOPS will be computed during training.
    
    \begin{table*}[t]
    	\centering
    	\footnotesize
    	\setlength\tabcolsep{0pt}
    	\begin{tabular*}{\textwidth}{@{\extracolsep{\fill}} *{12}{c} }
    		
    		\toprule 
    		{\textbf{Training}} & \multirow{2}{*}{\textbf{Model}} & \multirow{2}{*}{\textbf{Configuration}} & \multirow{2}{*}{\textbf{Layers}} & \multirow{2}{*}{\textbf{IPL}} & \multicolumn{5}{c}{\textbf{CS Ratio}} & \textbf{Avg.} & \multirow{2}{*}{\textbf{Parameters}} \\
    		\textbf{Data} \% &  &  & & & 10\% & 20\% & 30\% & 40\% & 50\% & \textbf{Score} \\
    		\midrule
    		
    		\multirow{4}{*}{100 \%}
    		& \multirow{2}{*}{ISTA-Net+ \cite{ista-net}}
    		& Baseline \cite{ista-net}   & 9  & 1 & 25.18/0.6971 & 28.11/0.8161 & 30.21/0.8769 & 32.12/0.9161 & 33.94/0.9421 & 29.91/0.8497 & 336,978 \\ 
    		& & Recursive                & 3  & 3 & \textbf{25.31}/\textbf{0.7010} & \textbf{28.21}/\textbf{0.8177} & \textbf{30.30}/\textbf{0.8786} & \textbf{32.21}/\textbf{0.9169} & \textbf{34.05}/\textbf{0.9428} & \textbf{30.02}/\textbf{0.8514} & \textbf{114,630}\\ \cmidrule{2-12}
    		
    		& \multirow{2}{*}{COAST \cite{coast2021}}
    		& Baseline \cite{coast2021}   & 20 & 1 & \textbf{26.24}/\textbf{0.7378} & \textbf{28.93}/\textbf{0.8378} & \textbf{30.99}/\textbf{0.8918} & \textbf{32.82}/\textbf{0.9255} & \textbf{34.58}/\textbf{0.9484} & \textbf{30.71}/\textbf{0.8683} & 1,122,024 \\
    		& & Recursive                 &  5 & 4 & 26.14/0.7329 & 28.79/0.8330 & 30.83/0.8879 & 32.68/0.9226 & 34.46/0.9463 & 30.58/0.8645 & \textbf{281,674}\\
			\midrule
			
			\multirow{4}{*}{3 \%}
			& \multirow{2}{*}{ISTA-Net+ \cite{ista-net}}
			& Baseline \cite{ista-net}   & 9  & 1 & 24.28/0.6614 & 27.06/0.7823 & 29.04/0.8498 & 30.87/0.8960  &  32.54/0.9264 & 28.76/0.8232 & 336,978 \\ 
			& & Recursive                & 3  & 3 & \textbf{24.71}/\textbf{0.6799} & \textbf{27.57}/\textbf{0.8017} & \textbf{29.65}/\textbf{0.8659} & \textbf{31.49}/\textbf{0.9066} & \textbf{33.20}/\textbf{0.9341} & \textbf{29.32}/\textbf{0.8376} & \textbf{114,630}\\ \cmidrule{2-12}
			
			& \multirow{2}{*}{COAST \cite{coast2021}}
			& Baseline \cite{coast2021}   & 20 & 1 & \textbf{25.88}/0.7226 & 28.55/0.8275 & \textbf{30.63}/\textbf{0.8848} & \textbf{32.44}/\textbf{0.9201} & \textbf{34.14}/\textbf{0.9440} & \textbf{30.33}/0.8598 & 1,122,024 \\
			& & Recursive                 &  5 & 4 & 25.87/\textbf{0.7252} & \textbf{28.57}/\textbf{0.8283} & 30.61/0.8846 & 32.41/0.9198 & 34.13/0.9438 & 30.32/\textbf{0.8603} & \textbf{281,674} \\
			\bottomrule
    		
    	\end{tabular*}
    	
    	\caption{PSNR/SSIM scores of the models on the BSD68 \cite{bsd68} dataset.}
    	\label{bsd68}
    \end{table*}

	\begin{table*}[t]
		\centering
		\footnotesize
		\setlength\tabcolsep{0pt}
		\begin{tabular*}{0.6\textwidth}{@{\extracolsep{\fill}} *{7}{c} }

            \toprule 
			\multirow{3}{*}{\textbf{Model}} & \multirow{3}{*}{\textbf{Configuration}} & \multirow{3}{*}{\textbf{Layers}} & \multirow{3}{*}{\textbf{IPL}} & \textbf{Training} & \textbf{Inference} & \multirow{3}{*}{\textbf{Parameters}} \\
                                            &                                         &                                  &                               & \textbf{Time}     & \textbf{Time}      & \\
                                            &                                         &                                  &                               & (Hours)           & (FPS)          & \\
			\midrule
			
			\multirow{2}{*}{ISTA-Net+ \cite{ista-net}}
			& Baseline \cite{ista-net} & 9  & 1 & 3.22 & \textbf{35.62} & 336,978 \\ 
			& Recursive                & 3  & 3 & \textbf{2.53} & 29.29 & \textbf{114,630}\\
            \midrule
			
			\multirow{2}{*}{COAST \cite{coast2021}}
			& Baseline \cite{coast2021}   & 20 & 1 & 18.14 & \textbf{12.27} & 1,122,024\\
			& Recursive                 &  5 & 4 & \textbf{10.50} & 12.20 & \textbf{281,674}\\
			\bottomrule
			
		\end{tabular*}
		
		\caption{Training time and number of parameters of the models in \cref{bsd68}. Note that the training time for ISTA-Net+ \cite{ista-net} was computed on a single CS ratio (10\%).}
		\label{stats}
	\end{table*}

\subsection{Recursion-based Feature Modulation Unit} \label{RFMU}
    To further tailor the model for recursions, we introduce a recursion-based feature modulation unit (RFMU).
    During training, it was observed that although naively adding recursions was sufficient to enhance the performance of the framework, it was not an optimal approach.
    Each phase cannot anticipate the level of quality of its input, and therefore, its features will not optimally enhance the results. Inspired by \cite{coast2021}, we extend the feature modulation unit seen in \cref{CPMB} by adding recursion statistics to $\mathbf{z}$. Specifically, the proposed RFMU unit will now modulate the features of the network based on two extra parameters. First, on how many iterations have been completed so far (denoted as $R_{\mathit{cur}}$), and second, the total number of iterations (which is $R_{\mathit{tot}} = \sum_i R_i$). For that end, we employ two linear layers separated by ReLU \cite{relu} that take the input vector $\mathbf{z}$ with the extra two parameters and extract $N_f$ modulation coefficients (denoted as $\sigma_{c}$). Finally, each feature in the output of the CPMB block will be modulated by its corresponding modulation coefficient $\sigma_{c}$. The figure of the RFMU unit is seen in \cref{fig:RFMU}.

    Originally, ISTA-Net+ \cite{ista-net} does not use a feature modulation unit, however, adding the RFMU to it is fairly simple and is shown in \cref{ISTAP_RFMU}. The only modification done to the network was the addition of the RFMU after the first convolutional layer and another RFMU before the last convolutional layer. Other than this and the inclusion of recursions, the recursive ISTA-Net+ model is identical to its reference in \cite{ista-net}.

\section{Experimental Results}
    For a fair comparison, all models were trained using the same dataset used in ISTA-Net+ \cite{ista-net} and COAST \cite{coast2021}, which features 88,912 blocks of dimension $33\times33$ obtained from 91 images. All models were trained using CS ratios from 10\% up to 50\%. The sampling matrix used was the random Gaussian matrix with orthogonal rows. For COAST however, we have generated multiple random Gaussian matrices for each CS ratio to faithfully follow their RPA training strategy \cite{coast2021}. The cost function was the mean-square error (MSE). The optimizer used was Adam \cite{adam} with the momentum set to 0.9, weight decay set to 0.999, and the learning rate set to $1\times10^{-4}$. The number of epochs was 400 for COAST, but for ISTA-Net+ \cite{ista-net}, it was trained for 200 epochs specifically as this is what was used in the original work \cite{ista-net}. All models were implemented using PyTorch \cite{pytorch}, and they were trained using a single Nvidia RTX 3090 GPU. For validation, we use the Set11 dataset \cite{reconnet} while for testing, we use the BSD68 dataset \cite{bsd68}. As for the evaluation metrics, we use the peak signal-to-noise ratio (PSNR) and structural similarity index (SSIM) \cite{SSIM}.

\subsection{COAST vs. Recursive COAST}
    To demonstrate the effectiveness of the recursive framework, we apply it to COAST \cite{coast2021}. The first model will use the same configuration as in \cite{coast2021} (labeled COAST). That is, it will have 20 layers without recursions. The second model will have 5 layers and 4 iterations per layer (labeled R-COAST).
    
    \cref{bsd68} shows the PSNR/SSIM scores for the BSD68 dataset. When both models have access to the entire training dataset, we observe that despite the recursive model only using 25\% the number of parameters of the baseline, it still achieves competitive results in terms of both the PSNR and SSIM metrics. However, when we limit the size of the training dataset to only 3\% of its original size, we find that the gap is mostly eliminated and both models achieve largely the same performance.
    
    Finally, although the recursive model sees a minimal impact on its performance, from \cref{stats}, we see that it takes 41\% less time to train and sees almost a 75\% reduction in its learnable parameters with a negligible impact on the inference time. Therefore, we can conclude that the recursive framework increased the efficiency of the COAST model with a minimal impact on its performance.

    \begin{figure} 
		\centering \includegraphics[width=0.65\columnwidth]{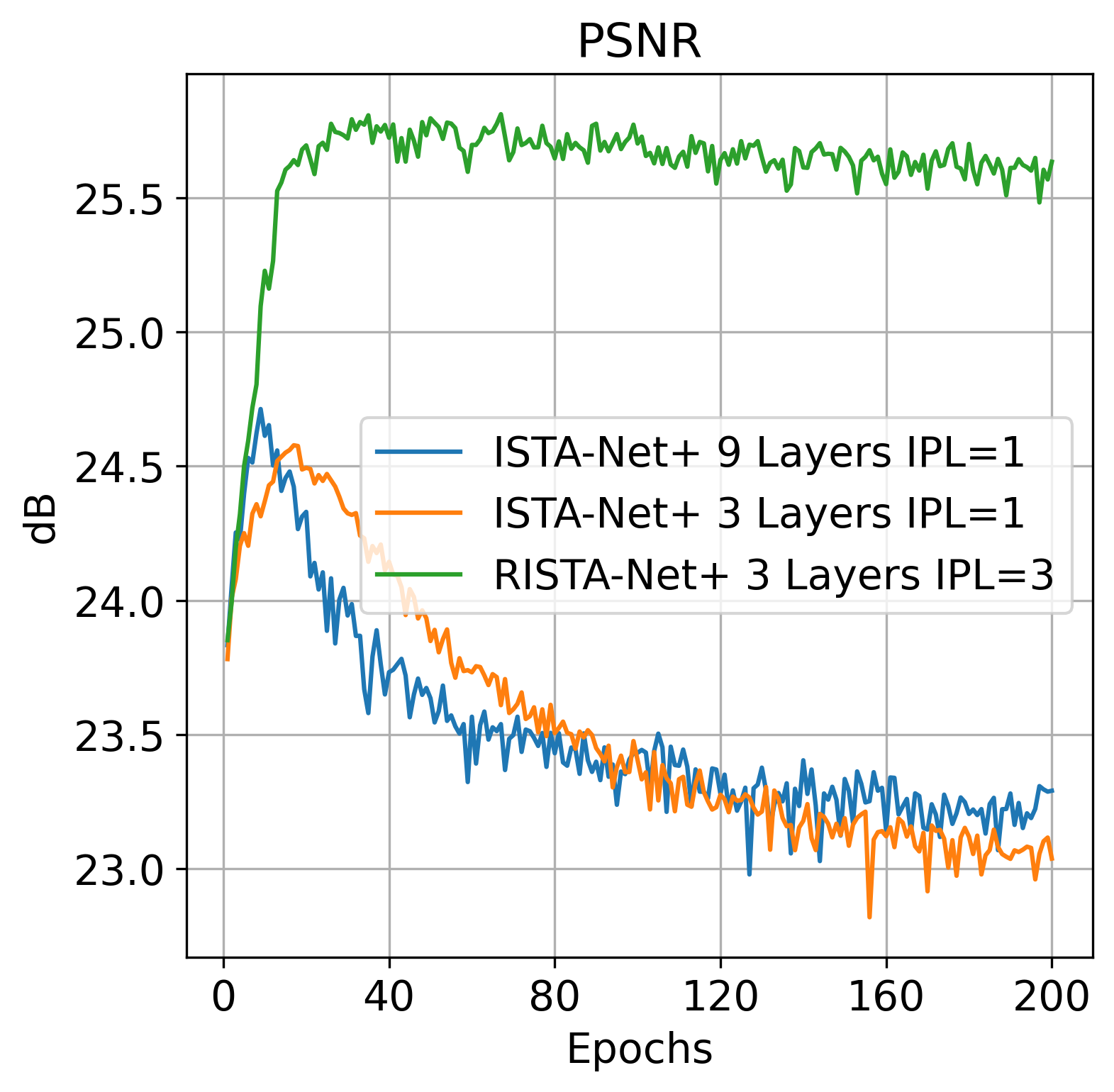}
		\caption{The learning curves of ISTA-Net+ \cite{ista-net} with and without recursions on the validation dataset (Set11 \cite{reconnet}). All models were trained using a CS ratio of 10\% and with access to only 3\% of the training dataset.
			\label{learning_curve_ista_lim}
		}
	\end{figure}

	\begin{figure}
		\centering
		\begin{subfigure}[b]{0.65\columnwidth}
			\centering
			\includegraphics[width=1\columnwidth]{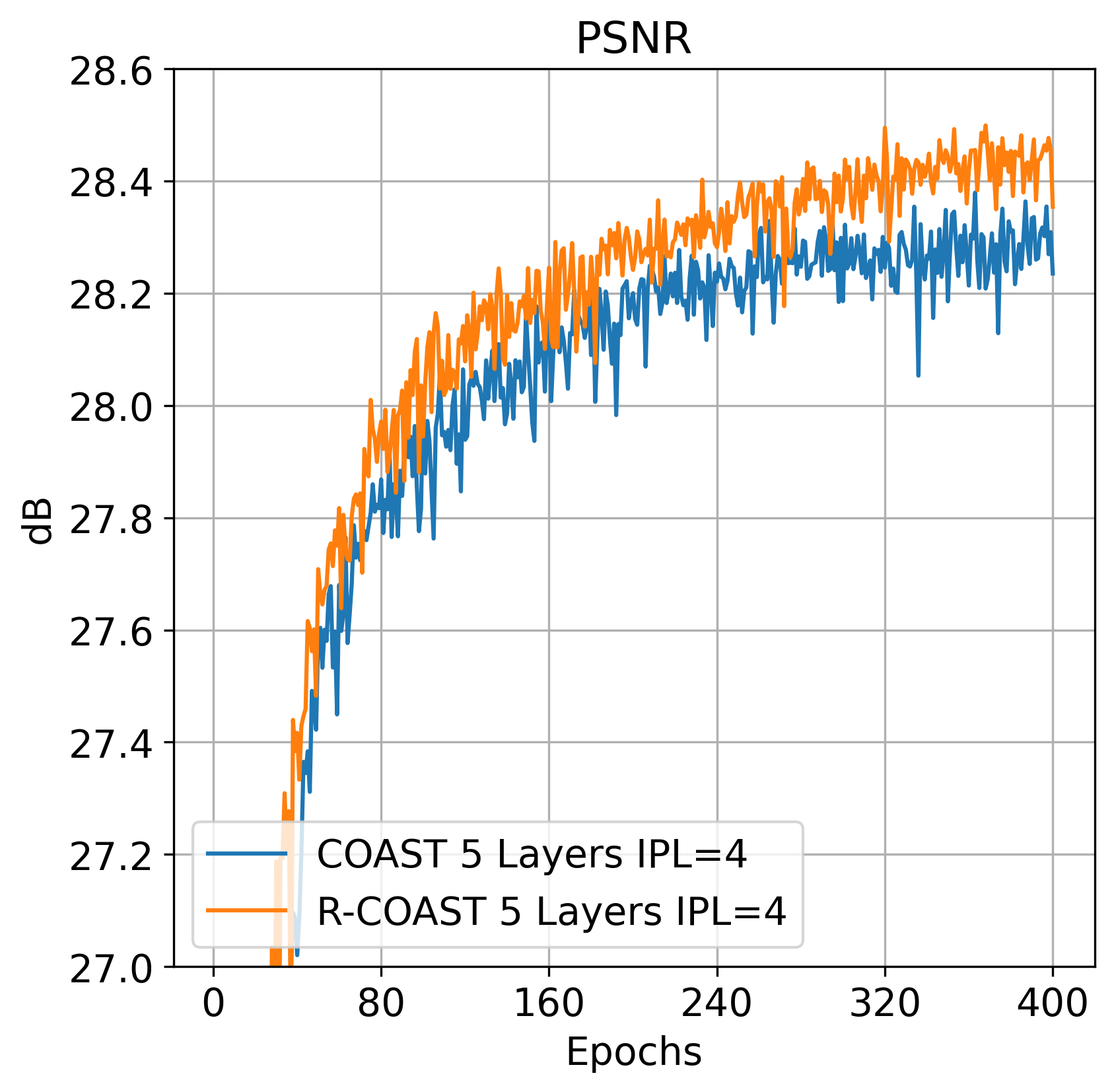}
			\caption{COAST \cite{coast2021} with and without the RFMU}
			\label{ablation_study_RFMU_COAST}
		\end{subfigure}
		\hfill
		\begin{subfigure}[b]{0.65\columnwidth}
			\centering
			\includegraphics[width=1\columnwidth]{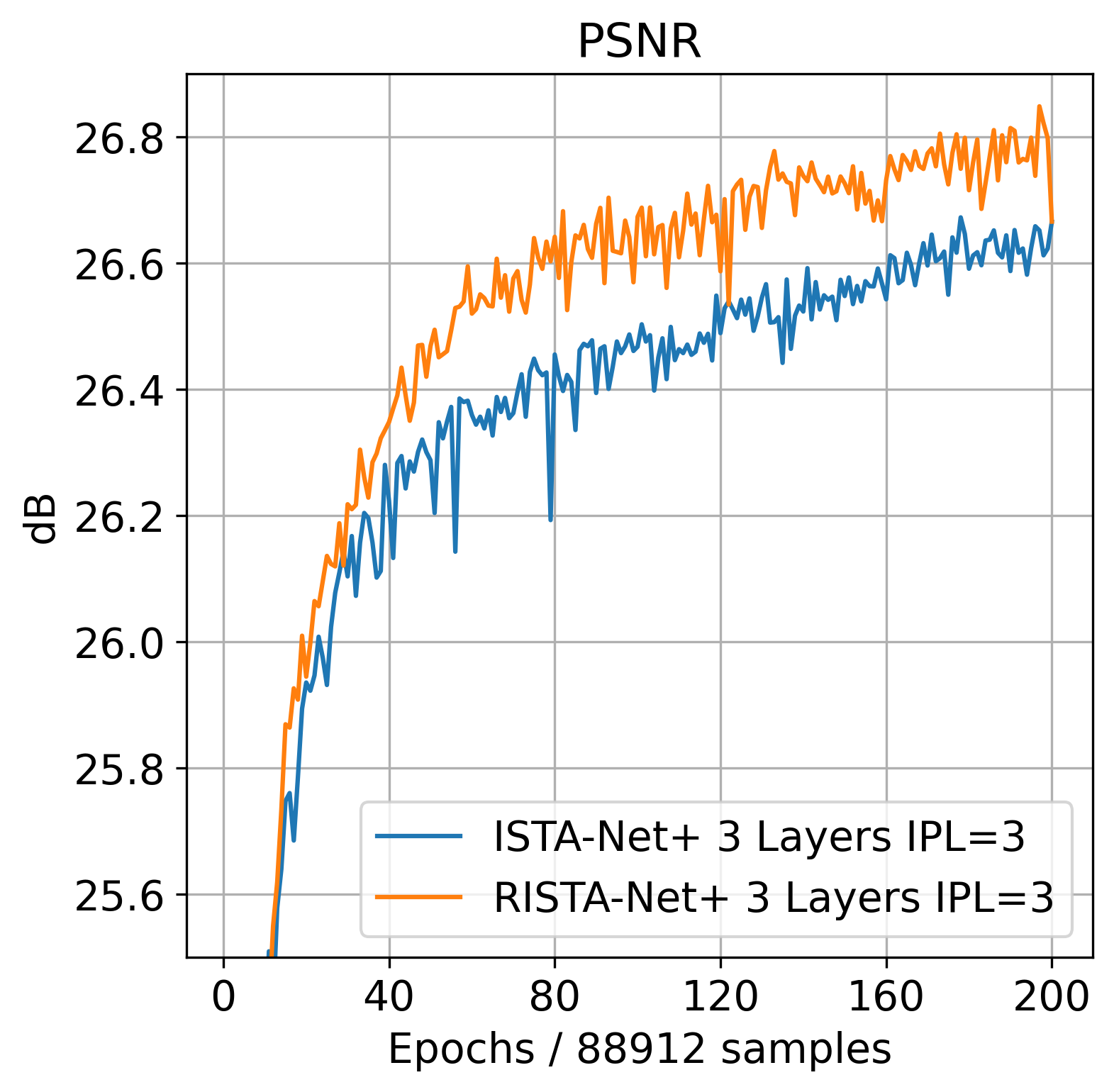}
			\caption{ISTA-Net+ \cite{ista-net} with and without the RFMU}
			\label{ablation_study_RFMU_ISTA}
		\end{subfigure}
		\caption{The learning curves of COAST \cite{coast2021} and ISTA-Net+ \cite{ista-net} with the RFMU (labeled R-COAST and RISTA-Net+ respectively) and without it (labeled COAST and ISTA-Net+ respectively) on the validation dataset (Set11 \cite{reconnet}) and at a CS ratio of 10\%.}
		\label{ablation_study_RFMU}
	\end{figure}

\subsection{ISTA-Net+ vs Recursive ISTA-Net+} \label{RISTA-Net}
	To investigate the generalizability of the recursive framework, we will compare the ISTA-Net+ \cite{ista-net} model with and without the recursive framework (labeled RISTA-Net+ and ISTA-Net+ respectively). We will train two models. One will use 9 layers with no recursions (the same configuration as in \cite{ista-net}), and the other model will use 3 layers and 3 IPL. The results are shown in \cref{bsd68}.
	
	On the testing dataset, we find that on average, the recursive configuration slightly outperforms the baseline model while using 66\% fewer parameters. However, when the training dataset is limited to only 3\% of its samples, we observe that the recursive configuration clearly outperforms the baseline. This is thanks to its inherently small number of parameters and therefore, it is more resilient to overfitting. This is also confirmed when looking at its learning curve in \cref{learning_curve_ista_lim}. Here, we observe that the recursive configuration is able to generalize better than the non-recursive model. We also include the results of a model with 3 layers (like RISTA-Net+) but without recursions, and the results in \cref{learning_curve_ista_lim} clearly show that the PSNR improvement RISTA-Net+ sees is \textit{not} attributed to its smaller size but rather its recursive design.
	
	\cref{stats} shows the training time and the number of parameters of the models. Here, it is observed that the recursive configuration takes 21\% less time during training thanks to the use of randomized recursions during training while only sustaining a minimal impact to the inference time.
	In general, we can conclude that applying the recursive framework to ISTA-Net+ \cite{ista-net} allows it to decrease its training time and number of parameters while maintaining the same performance when the training dataset is large, but when it is limited, the recursive model outperforms the baseline.

    \begin{figure*}[t]
		\centering
        \begin{subfigure}[t]{0.65\columnwidth}
			\centering
			\includegraphics[width=1\columnwidth]{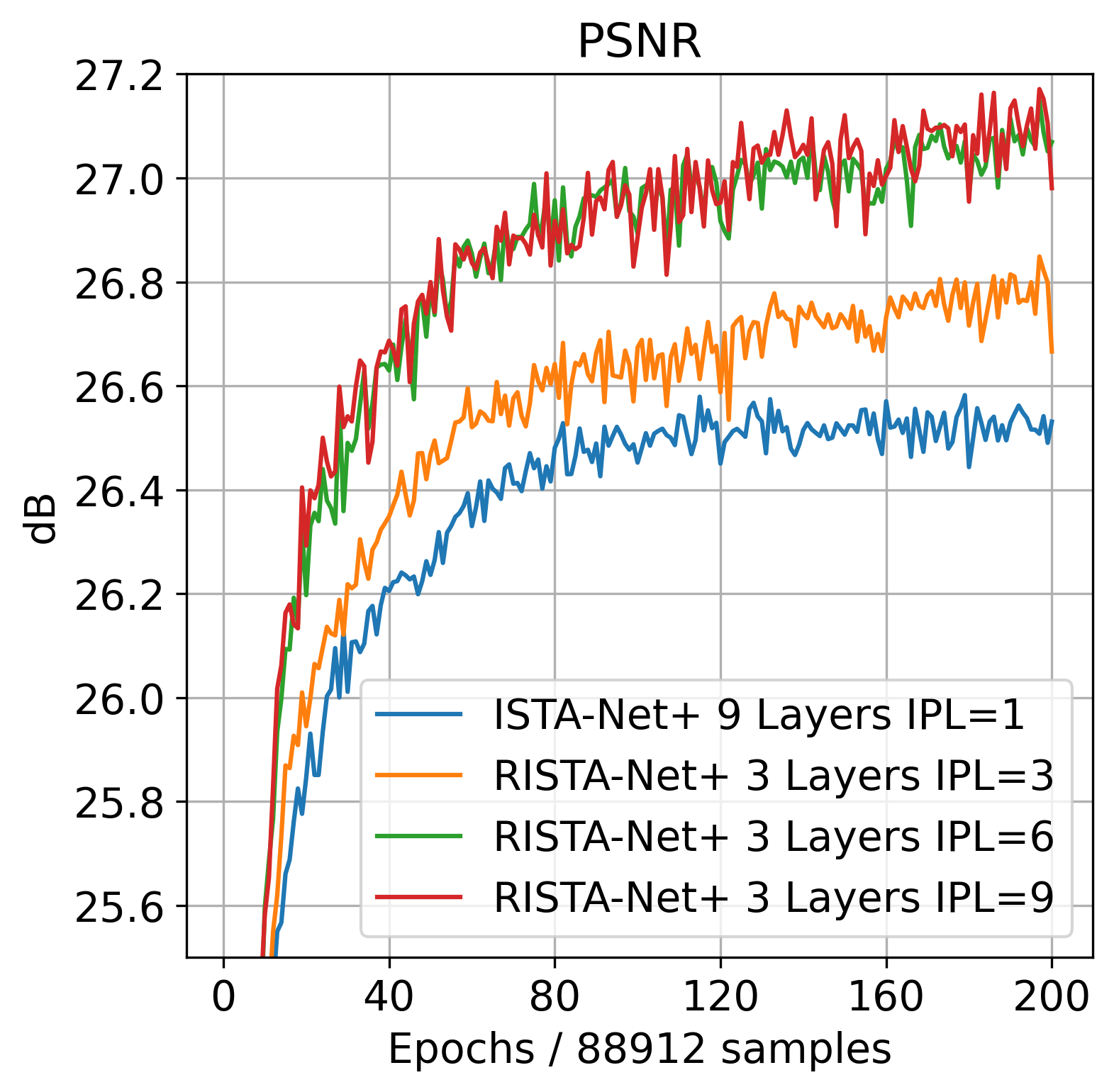}
			\caption{Learning curves of ISTA-Net+ \cite{ista-net}.}
			\label{ablation_study_recursion_limit_ista}
		\end{subfigure}
        \hfill
		\begin{subfigure}[t]{0.65\columnwidth}
			\centering
			\includegraphics[width=1\columnwidth]{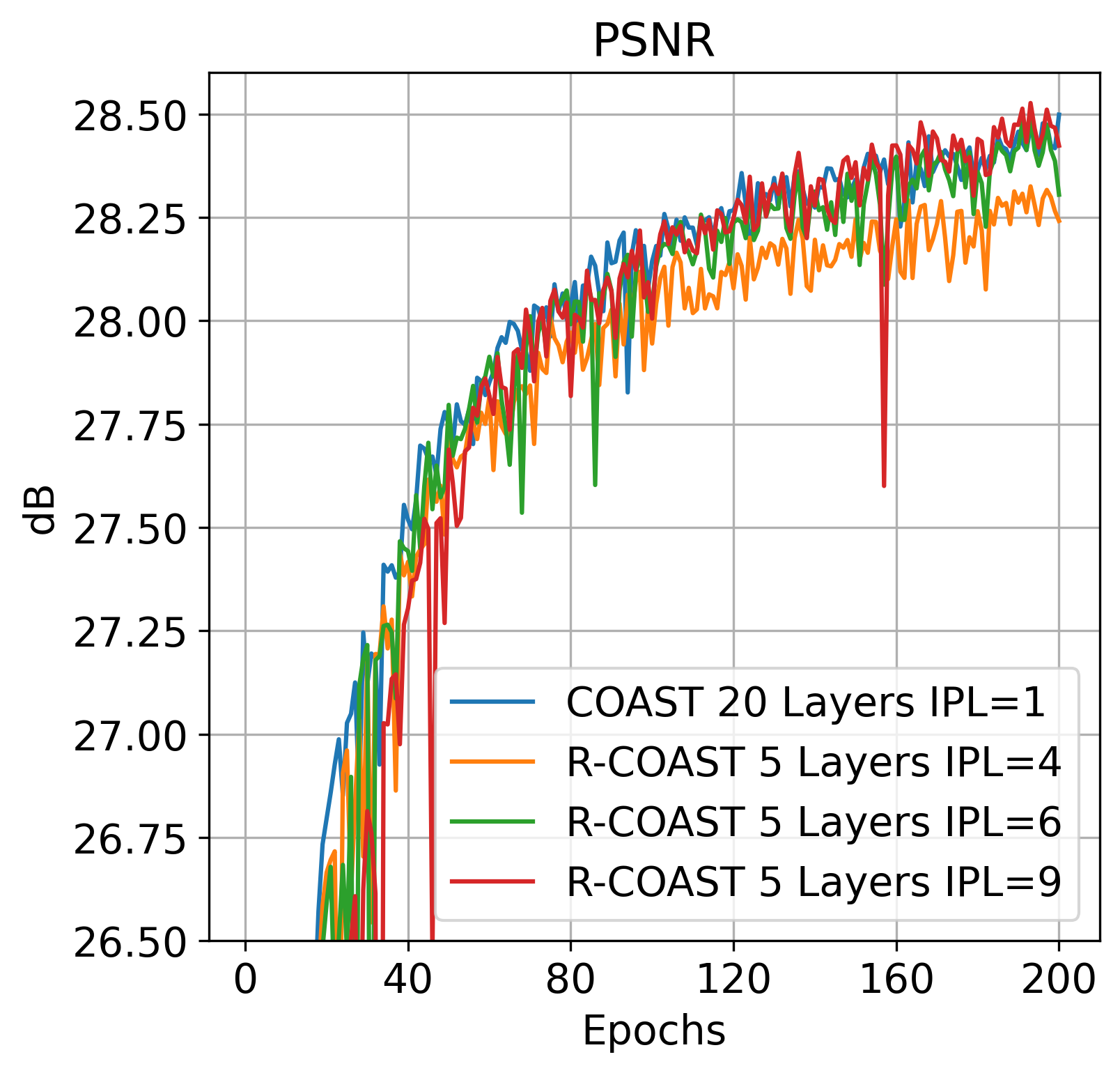}
			\caption{Learning curves of COAST \cite{coast2021}.}
			\label{ablation_study_recursion_limit_coast_raw}
		\end{subfigure}
		\hfill
		\begin{subfigure}[t]{0.65\columnwidth}
			\centering
			\includegraphics[width=1\columnwidth]{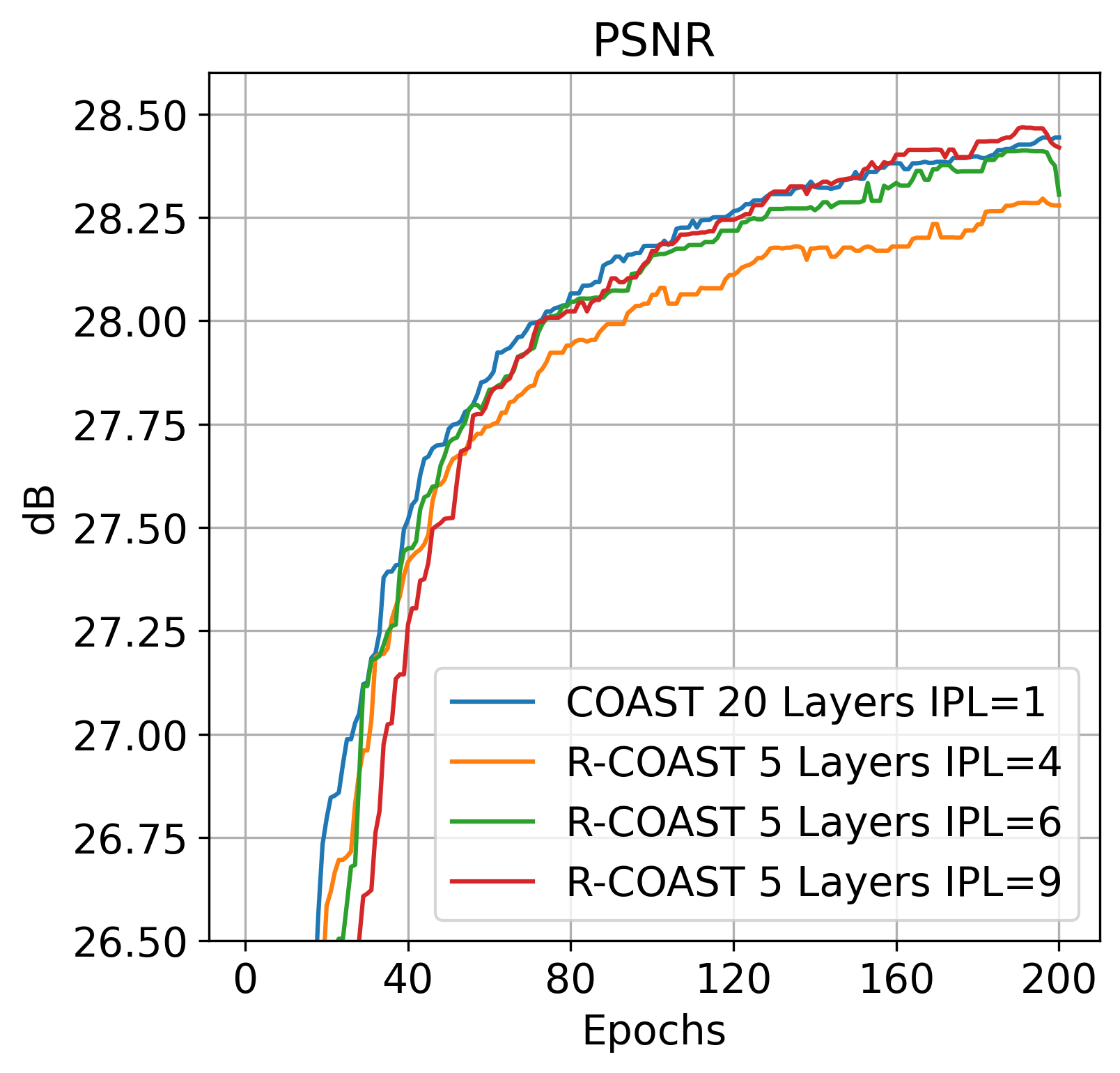}
			\caption{Learning curves of COAST \cite{coast2021} after passing through a median filter (to enhance visualization).}
			\label{ablation_study_recursion_limit_coast_median}
		\end{subfigure}
		\caption{The learning curves of ISTA-Net+ \cite{ista-net} and COAST \cite{coast2021} using different IPL values.}
		\label{ablation_study_recursion_limit}
	\end{figure*}

\section{Ablation Study}

\subsection{RFMU Enhances the Efficiency of Recursions}
    One of the problems in the naive implementation of recursions is that there can be a drastic difference in quality between each iteration (especially in the first iterations). More importantly, the model \textit{cannot} anticipate the level of quality of the input as it does not have access to the iteration index. Thus, it cannot modulate its features accordingly. In other words, the first iteration should have different weights than the last iteration, or at least some weights should be emphasized over the others. We proposed that modulating the features of the model based on recursion statistics improves its results. To investigate this, we compare the use of the RFMU in both COAST\cite{coast2021} and ISTA-Net+ \cite{ista-net}. The results are shown in \cref{ablation_study_RFMU}.

    \cref{ablation_study_RFMU_COAST} shows the results of COAST. Both models have 5 layers and 4 IPL, but one utilizes the RFMU (labeled R-COAST) while the other model does not (labeled COAST). Here, it is shown that the R-COAST model performs better as it achieved a higher PSNR score of around 28.50 dB while the COAST model without the RFMU scored 28.38 dB. From this, it can be concluded that the RFMU does indeed improve the performance of the model.
    
    The RFMU results of ISTA-Net+ are shown in \ref{ablation_study_RFMU_ISTA}. We trained both models to have 3 layers and 3 IPL, but again, one model foregoes the use of the RFMU (labeled ISTA-Net+) while the other model does (labeled RISTA-Net+). From \cref{ablation_study_RFMU_ISTA}, we observe that RISTA-Net+ can outperform ISTA-Net+ thanks to its use of the RFMU.
    
    Given the results in \cref{ablation_study_RFMU}, we can conclude that the addition of the RFMU allows the model to more efficiently utilize the recursive connection to achieve better results.

\subsection{Increasing the Number of Recursions}
    In the main results section, the maximum number of iterations per layer we performed was 4. The IPL was specifically chosen so that the total number of iterations for both the recursive model and the baseline remains the same. This raises the question: Can we see improved performance by increasing the IPL? And at what point would we experience diminishing returns?

    To investigate this, we trained both RISTA-Net+ and R-COAST using three different IPLs. For RISTA-Net+, all configurations used 3 layers and the chosen IPLs were 3, 6, and 9. For R-COAST, we settled on using 5 layers and the IPLs were 4, 6, and 9. The resulting learning curves in the validation dataset for both models are seen in \cref{ablation_study_recursion_limit} respectively.

    From the results in \cref{ablation_study_recursion_limit}, it is observed that both models see an increase in performance when the IPL is increased to 6. However, increasing the IPL to 9 sees a minimal increase in the performance, and thus, the tests indicate that 6 iterations per layer is the highest IPL possible before experiencing diminishing returns. Interestingly, when the R-COAST model uses 6 IPL, it actually closes the gap between it and the baseline model as it achieves roughly the same performance.

\section{Conclusion}
    In this work, we have discussed how most deep unfolding networks suffer from redundancies due to their iterative design. We argued that this can be mostly eliminated by using recursions. To demonstrate this, we applied our proposed framework to both COAST \cite{coast2021} and ISTA-Net+ \cite{ista-net}. In general, our findings confirm that the iterative design of such networks suffers from redundancies and that the use of the recursive framework successfully eliminates a lot of those redundancies. It can also decrease the training time by using randomized IPL during training. We believe that the most significant part of the recursive framework is its simplicity as this could allow it to be generalized to other networks since the core concept of the framework revolves around recursions, which are fairly simple to implement to other iterative networks. We believe that it is possible to extract even more utilization out of recursions as we have shown how implementing a simple block such as the RFMU can improve the efficiency of the framework. For future work, we would like to perform a more involved recursive framework and implement the recursions on the feature space instead of the spatial domain, as the increase of features could allow for more efficient use of recursions.
    
\section{Acknowledgement}
	Author(s) would like to acknowledge the support received from Saudi Data and AI Authority (SDAIA) and King Fahd University of Petroleum and Minerals (KFUPM) under SDAIA-KFUPM Joint Research Center for Artificial Intelligence.

\bibliographystyle{ieee_fullname}
\bibliography{refs}

\end{document}